\documentclass{article} % For LaTeX2e
\usepackage{iclr2023_conference,times}

\usepackage{amsmath,amsfonts,bm}

\def\eqref#1{equation~\ref{#1}}
\def\1{\bm{1}}

\def\mC{{\bm{C}}}

\DeclareMathAlphabet{\mathsfit}{\encodingdefault}{\sfdefault}{m}{sl}
\SetMathAlphabet{\mathsfit}{bold}{\encodingdefault}{\sfdefault}{bx}{n}

\usepackage{hyperref}
\usepackage{algorithm}
\usepackage{algpseudocode}
\usepackage{url}
\usepackage{graphicx}
\usepackage{multirow}

\title{Mask-Defense: An algorithm to reverse \\adversarial attacks on language models}
\title{Robustifying Language Models with Test-Time Adaptation}

\author{Noah T. McDermott, Junfeng Yang \& Chengzhi Mao  %\thanks{ Use footnote for providing further information
\\
Department of Computer Science\\
Columbia University\\
New York, NY 10027, USA \\
\texttt{\{ntm2128,cm3797\}@columbia.edu} \\
\texttt{junfeng.yang@cs.columbia.edu}
}

\iclrfinalcopy % Uncomment for camera-ready version, but NOT for submission.
\begin{document}

\maketitle
\begin{abstract}

% Recent research \citep{myfriendspaper} has found adversarial attacks on common image classification neural networks can be reversed by using knowledge gained from self-supervised learning algorithms to modify input images at test time.  We developed a novel strategy to defend against attacks in large-scale natural language models like BERT in a similar manner, leveraging information from masked-language objectives to defend against adversarial attacks through word substitution.  We were able to achieve high rates of success against state of the art adversarial attacks on datasets such as Imdb and AgNews without impacting accuracy on clean sentences.

Large-scale language models achieved state-of-the-art performance over a number of language tasks. However, they fail on adversarial language examples, which are sentences optimized to fool the language models but with similar semantic meanings for humans. While prior work focuses on making the language model robust at training time, retraining for robustness is often unrealistic for large-scale foundation models. Instead, we propose to make the language models robust at test time. By dynamically adapting the input sentence with predictions from masked words, we show that we can reverse many language adversarial attacks. Since our approach does not require any training, it works for novel tasks at test time and can adapt to novel adversarial corruptions. Visualizations and empirical results on two popular sentence classification datasets demonstrate that our method can repair adversarial language attacks over 65\% of the time. 

% but also it is compatible with all pretrained foundation models
\end{abstract}

\section{Introduction}
Large-scale pretrained language models (foundation models) like BERT \citet{bert} and RoBERTa \citep{roberta} have achieved state of the art performances over a number of language tasks, such as sentiment classification and completion~\citep{bert_perf}.  However, these models are vulnerable to adversarial attacks, where modifications to inputs, imperceptible to humans, cause machine learning models to misclassify.  These vulnerabilities can pose a security risk when they are used in sensitive and safe-critical applications~\citep{realworld}.

Existing defenses against adversarial attacks have largely focused on training, either through training on pre-generated adversarial samples~\citep{dataapproaches, mao2022understanding}, or modifying the training objective to be more robust~\citep{modeldef}.  However, this is an inherently difficult task, as there are a vast number of different types of attacks on the character, word, and sentence level that the model would have to be able to defend against, and it cannot adapt to new, novel types of attacks~\citep{limitations}. In addition, training-based approaches can only achieve robustness on the task that they have been trained on, but cannot generalize to novel tasks, which is a key feature for modern language foundation models~\citep{limitations}.

Our approach shifts the burden of robustness from training to test time.  Our key insight is that masked language modeling is able to capture the structure and constraints of a natural language sentence which are violated in adversarial attacks.  We use masked language modelling, a self-supervised task, to find key words in adversarial sentences that lead to bad predictions, and replaces them with normal words. 

Our adaptation algorithm for robustness not only can achieve robustness on novel adversarial attacks, but also can achieve robustness in a zero-shot manner without performing robust training on the downstream tasks. Since our method is at test-time, it is also compatible with all existing training-based robust algorithms.

% We also ensure that the new words are synonymous with the replaced words to ensure the sentence meaning remains the same.

We ran experiments against two of the latest text-based adversarial attacks, PWWS \citep{pwws} and TextFooler \citep{tfpaper}.  Empirically, our experiments show that our defense, called Mask-Defense, was able to reverse 75-80\% percent of successful Textfooler attacks, and 65-70\% successful PWWS attacks.  Additionally, our defense continues to correctly classify sentences that have not been attacked, with around 98\% of correctly classified clean sentences remaining correct after the defense is run.  The reverse attacked sentences also display high levels of semantic similarity with the original sentences.  

\begin{figure}[h]
\begin{center}
\label{Summaryfig}
%\framebox[4.0in]{$\;$}
%\fbox{\rule[-.5cm]{0cm}{4cm} \rule[-.5cm]{4cm}{0cm}}
\includegraphics[width=\textwidth]{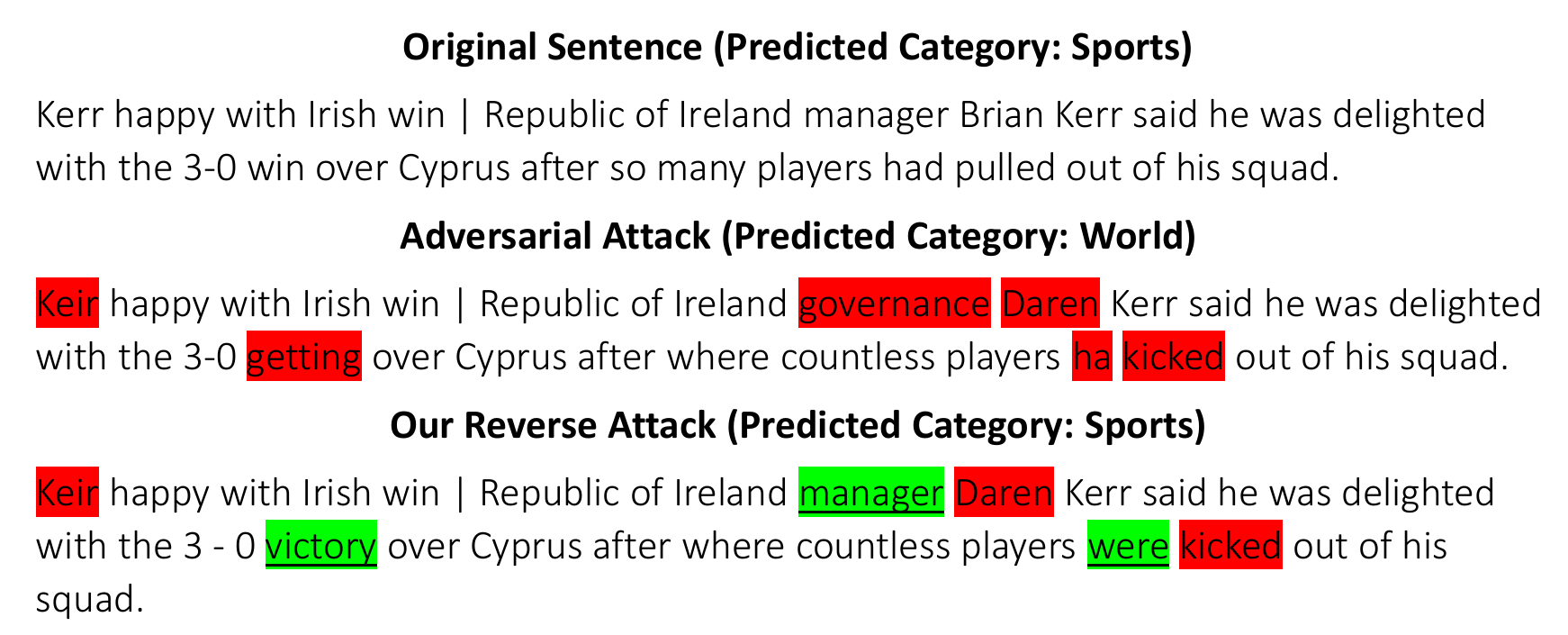}
\end{center}
\caption{An example of an adversarial attack successfully reversed by Mask-Defense. We show the original sentence from the Ag's News dataset, and the adversarial attacked sentence using TextFooler (attacked words in red). Reverse Attack shows the adapted sentence using our algorithm, which locates high-loss words and replaces them (green words) with examples generated from a Masked-Language model. }
\end{figure}
 \section{Related Work}
 \subsection{Text-based Adversarial Attacks and Defenses}

 A large number of adversarial attacks have been proposed against language models.  Unlike those commonly used in vision or speech, there is no way to imperceptibly add noise to text.  So instead, attacks focus on making changes at the character \citep{hotflip}, word \citep{pwws} \citep{tfpaper}, or sentence\citep{t3attack} level that keep the sentence's meaning intact, but creating a different output.

 Various defenses against these text-based attacks have been proposed, but most are training-based approaches, either through adding adversarial attacks in a form of data augmentation, \citep{dataapproaches} or by modifying the training goal to improve robustness.\citep{modeldef}  While these approaches have been empirically successful, they are difficult to implement because of the computation needed to generate enough adversarial attacks and because of the vast number of possible attacks that the network must be trained to defend against.
 
\subsection{Defenses Leveraging Self-Supervised Learning}
Self-supervised learning originated as a way to solve the problem of expensive labeled data.  Since complex machine learning models often require large amounts of data, and that data must be manually labeled by humans, it often becomes a bottleneck in development.  Self-supervised learning solves that problem by teaching a machine learning model to predict one part of its input using the rest of its input.  It can be used as a means to an end, or it can be used for transfer learning with a relatively small amount of labelled data.  One of the most successful examples is BERT\citep{bert} which is trained on masked-language modelling and next-sentence prediction, both self-supervised methods, but also can achieve strong results on other tasks such as sentiment analysis when fine tuned with small amounts of labelled data.

\citet{myfriendspaper} developed an algorithm to perform a "reverse attack" on image classifier models at inference time using self-supervised learning.  The algorithm uses contrastive loss as a self-supervised objective, takes the gradient of that loss, and adds a perturbation in the direction of that gradient.  By doing this, it is able to restore some of the natural structure in the image, and it achieves strong performance against a number of attacks. \citet{lawhon2022using} shows multiple tasks further improves the robustness.
\citet{mao2022robust, zhang2022adversarially} generalizes this line of work to image segmentation and video perception.
This line of work~\cite{tsai2023selfsupervised} all requires multiple steps of test-time optimization which involves the gradient descent. 
In contrast, our work is tailored for language domain, which replace the words with masked token prediction only requires feed-forward pass, which is much faster to implement.

% Another advantage of this kind of defense is that it requires no type of specialized adversarial training and is done completely at inference time, allowing it to be combines with other types of defenses such as adversarial training.

\section{Method: Robustifying Language Models via Masked Word Prediction}
% \subsection{Preliminaries}
\textbf{Attacks.}
From the definition given by \citet{tfpaper}, given a corpus of sentences $\mathcal{X} = \{ X_1, X_2, X_3 \dots \}$ and a set of output labels $\mathcal{Y} = \{ y_1, y_2, y_3 \dots \}$, a pre-trained model $F$ performs some mapping $F: \mathcal{X} \longrightarrow \mathcal{Y}$.  A valid adversarial attack $X_{adv}$ on $X \in \mathcal{X}$ is some sentence that satisfies the properties

$$F(X) \neq F(X_{adv}) \textnormal{  and  } S(X, X_{adv}) \geq \epsilon$$ 

Where $S(A, B) \longrightarrow [0, 1]$ is a similarity metric that returns a higher value when two sentences $A, B \in \mathcal{X}$ are similar in meaning and structure.

\textbf{Reverse Attacks.}
The goal of our algorithm is to reverse the attack at inference by creating some new sentence $X_{def}$ from $X_{adv}$ such that:

$$F(X_{def}) = F(X) \neq F(X_{adv}) \textnormal{ and } S(X_{def}, X_{adv}) \geq \epsilon$$

The main difference between the reverse attack setting and the attack setting is that the defender does not have access to output of $F$ or to the output of any similarity metrics, and this makes it much more challenging than the attack setting.  However, the defender can use self-supervised learning to gain information without accessing the outputs, and this strategy is central to our algorithm.

% \subsection{Our Algorithm: Mask-Defense}
\begin{algorithm}
\caption{Mask-Defense}\label{alg:cap}
\begin{algorithmic}
\State \textbf{Inputs: } Sentence $S$, Classifier $F$, Masked-Language Model $M$, Cosine Similarity Matrix $C$, Model Vocabulary $V_M$, Matrix Vocabulary $V_C$, Threshold $\alpha$, Replacement Parameter $n$.
\State \textbf{Outputs: } New sentence $\hat{S}$
\State $\hat{S} \gets S$
\For{$i \in \{0, \textnormal{len}(S)\}$}
\State Let $\hat{S}_i$ be sentence $S$ with the $i$-th word masked out
\State Use $M$ with $\hat{S}_i$ as an input to calculate the masked-language modelling loss $\mathcal{L}_i$ and the softmax output $o_i$ over $V_M$
\EndFor
\State $j \gets 1$, $r \gets 0$
\While{$r < n$ and $j \leq 50$}
\State Let $i$ be the word position that has the $j$-th highest masked-language modelling loss $\mathcal{L}_i$
\For {$k \in \{0, 50\}$}
\State Let $w_i$ be the word in $S$ at position $i$
\State Let $o_{i}^k$ be the word in the vocabulary with the $k$-th highest softmax output in $o_i$
\If {$\displaystyle \mC(w^i, o_i^k) \geq \mu(\displaystyle \mC) + \alpha \sigma(\displaystyle \mC)$ and $w_i \in V_C$ and $o_i^k \in V_C$}
\State Replace $w_i$ with $o_i^k$ in $\hat{S}$
\State $r=r+1$
\State \textbf{break}
\EndIf
\EndFor
\State $j=j+1$
\EndWhile
\end{algorithmic}
\end{algorithm}

The self-supervised task we choose to optimize for is called masked-language modelling.  Masked language modelling consists of taking a sentence, masking some of the words, and having the model choose the best words that fit in its vocabulary.  The loss is taken as a softmax over the all the words in the vocabulary.

The algorithm has two hyperparameters that can be adjusted depending on how the user wants to balance success rate and similarity.  These are $n$, the total number of word replacements, and $\alpha$, the minimum word similarity score.  Replacing a fraction of words in a sentence instead of a fixed number was considered for longer sentences, but experiments showed a diminishing effect. 

\textbf{Step 1: Word Importance Ranking}
We begin this algorithm by ranking the words in the sentence by importance.  However, we do not have access to the classifier model's outputs, so we instead use masked-language modelling loss as a measure of importance.  For each word $w$ in some sentence $X$, we calculate the importance $I_w$ as: $$I_w = L(F(X_w))$$  Where $X_w$ is the sentence $X$ with word $w$ masked, and $L$ is the cross-entropy loss taken over the softmax of the entire vocabulary.  The justification for using this score is simple: A masked word that only has a few potential candidates will end having a low importance, such as "I saw [Mask] big rabbit.".  Clearly "a" is one of the only words that would fit here.  On the other hand, "I saw a big [Mask]." could have many valid words and therefore will have a high importance and will be a good candidate for substitution.  While this approach may not always be accurate, such as in the case of an unexpected word shows up even in when there are few valid word candidates, our experiments show that on average it is a good measure of importance when there is no access to the classifier model.

This is the most computationally complex part of the algorithm, as $I_w$ must be calculated for every word $w \in X$ and sorted.  In addition, to calculating the losses, the highest 50 logits $l_w$ (outputs from before the softmax layer) are also saved.  Word replacement is then applied from highest to lowest importance.

\textbf{Step 2: Word Replacement}
Next, begin the process of replacing words in the sentence with MLM predictions using the logits.  However, MLM predictions only capture context and not meaning, for example, an MLM model may predict either "good" or "bad" for the sentence "Today was a [Mask] day.", so we introduce an additional model to check word meaning.  We use word embeddings from \citet{counterfittedembs}, which were designed to capture synonymy.

To determine whether two words are similar, we build a cosine similarity matrix $\displaystyle \mC$ over the entire embeddings, and take the distance between the original and new word.  The threshold to determine whether a new word $n$ is similar to some original word $o$ is: $$\displaystyle \mC_{o, n} \geq \mu(\displaystyle \mC) + \alpha \sigma(\displaystyle \mC)$$

Where $\alpha$ is the similarity parameter, and $\mu$ and $\sigma$ are the mean and standard deviation across all elements in the matrix.  The algorithm considers each MLM word candidate in order of likelihood, and chooses the first one with a similarity score above the threshold.  The vocabulary of the MLM model and the embeddings may differ, so any word that is in the MLM vocabulary but not the embeddings' vocabulary is not considered.  If no successful candidates can be found in the top 50 predictions, or the MLM model predicts the original word, then that word is not replaced.  The algorithm terminates when $n$ words in total are replaced. 

\begin{figure}[h]
\begin{center}
\label{Summaryfig}
%\framebox[4.0in]{$\;$}
%\fbox{\rule[-.5cm]{0cm}{4cm} \rule[-.5cm]{4cm}{0cm}}
\vspace{-10mm}
\includegraphics[scale=.6]{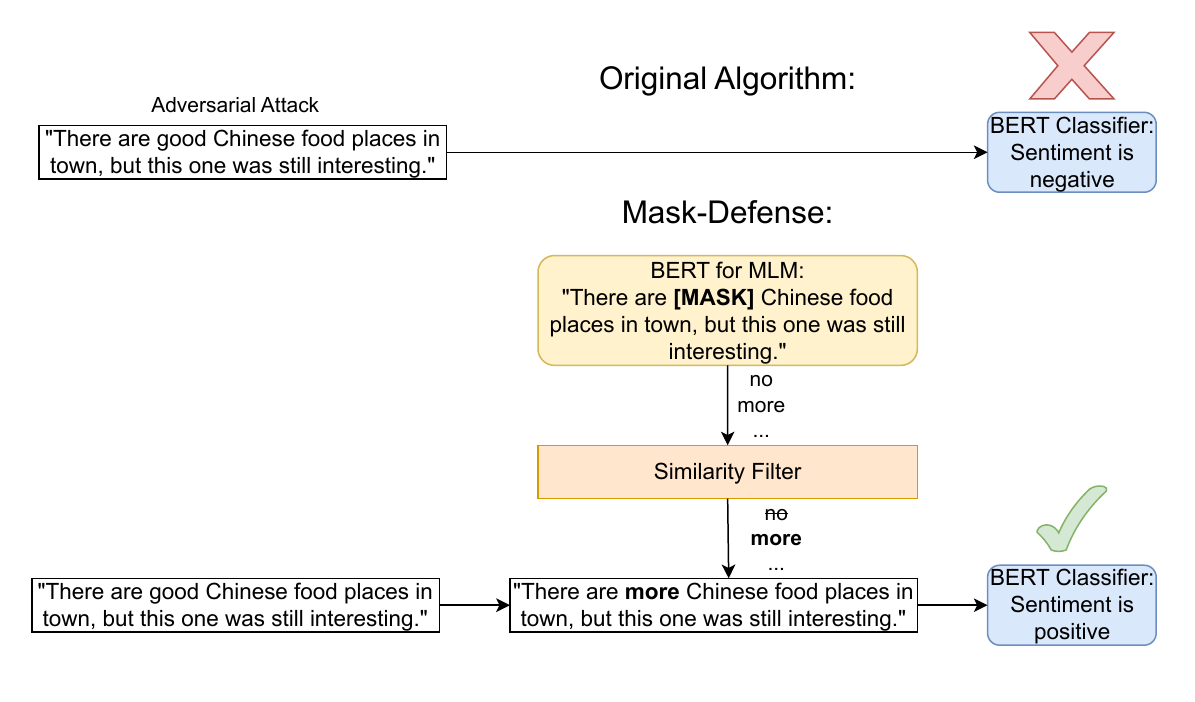}
\end{center}
\vspace{-5mm}
\caption{An overview of the Mask-Defense algorithm.  A masked-language model generates new words, which are filtered by semantic similarity, to modify input sentences to a language model at test time.}
\end{figure}
\section{Experiments}

We evaluate this defense on two state-of-the-art attacks on different text classification tasks.  To generate the adversarial text, we run the specified attack against the model until we have 1000 successful examples, and then we run our defenses against each.  Unsuccessful attacks were not considered.  In addition, we also ran the attack against 1000 clean sentences from both datasets to ensure that the defense does not change the output on the clean sentences.

\textbf{Datasets}
The two datasets used in this experiment are AG's News and Yelp Polarity.  AG's news is a sentence level classification dataset of the title and description of news stories into four categories: Word, Sports, Business, and Science/Technology.  Yelp is a document-level sentiment classification dataset of reviews of restaurants, businesses, etc. Reviews with 1 or 2 stars are considered negative, while reviews with 4 or 5 stars are considered positive.  Classification, attacks, and defenses were done by using the entire document as an input, as opposed to individual sentences of the review.

\textbf{Models.}
The model being attacked are is BERT \citep{bert} a common baseline in the NLP world that has been very successful in both self-supervised and downstream tasks.  BERT is also used to calculate the masked-language-modelling loss in the attack.  For classification, fine-tuned models from the Textattack package were used, which used hyperparameter optimization to get the best results on Yelp \footnote{https://huggingface.co/textattack/bert-base-uncased-yelp-polarity} and Ag's News \footnote{https://huggingface.co/textattack/bert-base-uncased-ag-news}.  The standard bert-base-uncased model was used to generate MLM tokens.
\begin{table}[t]
\renewcommand{\arraystretch}{1.25}
\label{results-table}
\begin{center}
\setlength\tabcolsep{2.5pt} 
\begin{tabular}{|c|c|c|c|c|}
\hline
Dataset & Attack Method & Clean Classification Unsuccessful & Attack Unsuccessful & Attack Successful \\
\hline
 \multirow{2}{*}{Ag's News} & Textfooler & 5.6\% & 19.8\% & 74.6\% \\
 & PWWS & 5.1\% & 41.2\% & 53.1\% \\
 \hline
\multirow{2}{*}{Yelp} & Textfooler & 3.2\% & 5.6\% & 91.1\% \\
& PWWS & 3.2\% & 5.7\% & 91.2\% \\
\hline

\end{tabular}
\caption{Analysis of Attack Efficiency. 
 The classifier models used in this experiment had a high rate of success againt clean sentences, but were also able to be defeated by adversarial attacks.}\label{tab:analysis_attack}
\vspace{-5mm}
\end{center}
\end{table}

\textbf{Attacks and Defense.}
The two attacks the algorithm defends against are PWWS \citep{pwws} and TextFooler \citep{tfpaper}.  PWWS is an early baseline for NLP adversarial attacks, which performs word replacement combining word saliency and classification probability.  Textfooler searches for similar words in embeddings, and uses part-of-speech tagging and sentence encoders to ensure similarity to the original sentence.  Both attacks were implemented with Textattack, a python package for adversarial attacks, and were run with default hyperparameters.\footnote{https://github.com/QData/TextAttack}.  

The defense was run with hyperparameters of $\alpha = 2$ and $n = 3$, meaning that words must have a similarity score of two standard deviations above the mean to be replaced, and at most three words in one input may be replaced.  Only successfully classified sentences were attacked, and the defense was only run on successfully attacked sentences and clean examples.

\begin{table}[t]
\renewcommand{\arraystretch}{1.25}
\label{results-table}
\begin{center}
\setlength\tabcolsep{2.5pt} 
\begin{tabular}{|c|c|cc|cc|cc|}
\hline
&&\multicolumn{2}{|c|}{Clean} & \multicolumn{2}{|c|}{TextFooler}  &\multicolumn{2}{|c|}{PWWS} \\ 
Dataset & Defense & Accuracy & Similarity & Accuracy & Similarity & Accuracy & Similarity \\
\hline
\multirow{2}{*}{Ag's News}  & Baseline & 1 & n/a & 0 & n/a & 0 & n/a \\
& \textbf{Mask-Defense} & \textbf{0.982} & \textbf{0.948} & \textbf{0.790} & \textbf{0.938} & \textbf{0.688} & \textbf{0.936}\\
\hline
\multirow{2}{*}{Yelp}  & Baseline & 1 & n/a & 0 & n/a & 0 & n/a \\
& \textbf{Mask-Defense} & \textbf{0.987} & \textbf{0.950} & \textbf{0.756} & \textbf{0.958} & \textbf{0.657} & \textbf{0.952}\\

\hline
\end{tabular}
\caption{Experimental Results.  75-80\% of successful Textfooler attacks and 65-70\% of successful PWWS attacks were reversed by Mask-Defense.  98-99\% of correctly classified clean sentences remained correct after Mask-Defense.  New sentences created by Mask-Defense had a very high similarity to the original ones.}\label{tab:exp} \vspace{-5mm}
\end{center}
\end{table}

\textbf{Results}
We show the model accuracy under popular attack in Table \ref{tab:analysis_attack}. Before the adversarial attack, the prediction failure rate is 3-5\%. The attack will only focus on fool the examples that are successfully classified. After applying adversarial attack, it can further subvert 53\%-91\% of the samples in the total test set. The established attack can fool the language classifier. In the following experiment, we will focus on correcting the examples that are subverted by the attacker.

We show the robust accuracy of our defense algorithm in Table~\ref{tab:exp}. Mask-Defense is able to achieve a very high success rate with very few modifications to the text.  Interestingly, the average sentence length in Yelp (179.18 words) was over three times that of Ag's News (53.17) yet the results were very similar even though in both cases only three words were allowed to be replaced.  This suggests that the replacement of a few words with high MLM loss was enough to reverse the adversarial attack regardless of sentence length.  Even though there was no sentence-level bound on similarity, all defenses achieve a similarity score of 0.93 or higher.  Therefore, it can be assumed that switching a small amount of words in a sentence with very similar replacements is a good proxy for sentence level similarity.  Another interesting observation is the fact that PWWS attacks were more difficult to reverse despite the fact that PWWS is an older and less sophisticated algorithm.  This may be due to the tighter constraints on syntactic and semantic similarity that Textfooler requires.  Finally, we also see an almost perfect success rate on the clean sentences, which shows that Mask-Defense does not negatively impact sentences which have not been attacked.

\begin{table}[t]
\renewcommand{\arraystretch}{1.25}
\label{analysis-table}
\begin{center}
\setlength\tabcolsep{2.5pt} 
\begin{tabular}{|c|c|cc|}
\hline
&&\multicolumn{2}{|c|}{Average Loss Before Reverse} \\ 
Dataset & Outcome & TextFooler & PWWS\\
\hline
\multirow{2}{*}{Ag's News}  & Success & 1.28 & 1.58\\
& Failure & 2.24 & 2.75\\
\hline
\multirow{2}{*}{Yelp} & Success & 1.47 & 2.56 \\
& Failure & 2.69 & 3.58\\

\hline
\end{tabular}
\caption{Analysis of failure.  Attacks that were successfully reversed had a much smaller cross-entropy loss from the classifier model, showing that mask-defense performed much better on attacks that were on the edge of the decision boundary}\label{tab:analysis}
\vspace{-5mm}
\end{center}
\end{table}

\textbf{Analysis.} Table \ref{tab:analysis} provides some insight on why mask-defense failed in some instances.  For all attack methods on all datasets, attacks that were successfully reversed had a much lower cross-entropy loss from the classifier to begin with.  This highlights how the defense setting is significantly more difficult than the attack setting; while attack models like TextFooler can continue to change words until the label changes, Mask-Defense has no access to the output labels and can only change a pre-defined number of words.  For attacks that are right on the decision boundary, this approach works well, but for more severe or perturbed examples, it falls short.  A possible improvement to Mask-Defense would be to quantify how perturbed some sentence is, and only change certain words.

We use sentence similarity to measure how close our generated sentence to the original sentence. The similarity is calculated by using a universal sentence encoder, a technique pioneered in \citet{use}.  We use it to encode the adversarial sentence and the new sentence into a 348-dimension space, and use their cosine similarity as a measurement of similarity.  The exact pretrained model used in our experiments experiments can be found here\footnote{https://huggingface.co/sentence-transformers/all-MiniLM-L6-v2}.
In Table~\ref{tab:exp}, we show that our method can fix the adversarial sentence by making them more similar to the original clean sentence.

\section{Conclusion}
We present a new algorithm that can use the underlying knowledge from masked-language-modelling to reverse state of the art textual adversarial attacks and restore ground truth.  The experiments show that this algorithm was able to achieve noteworthy results even with very strict limits on the changes it could make to the sentence, and that defenses used at test time in language models can be a powerful way to increase robustness.  

\section{Acknowledgement}
This research is based on work partially supported by the NSF NRI Award \#2132519, a GE/DARPA grant, a CAIT grant, and gifts from JP Morgan, DiDi, and Accenture.

\bibliography{iclr2023_conference}
\bibliographystyle{iclr2023_conference}

\end{document}